# Reverse Queries in DATR*

Hagen Langer

University of Osnabrück, Germany
hlanger@jupiter.rz.uni-osnabrueck.de

## Abstract

DATR is a declarative representation language for lexical information and as such, in principle, neutral with respect to particular processing strategies. Previous DATR compiler/interpreter systems support only one access strategy that closely resembles the set of inference rules of the procedural semantics of DATR (Evans & Gazdar 1989a). In this paper we present an alternative access strategy (*reverse query strategy*) for a nontrivial subset of DATR.

## 1 The Reverse Query Problem

DATR (Evans & Gazdar 1989a) has become one of the most widely used formal languages for the representation of lexical information. DATR applications have been developed for a wide variety of languages (including English, Japanese, Kikuyu, Arabic, Latin, and others) and many different subdomains of lexical representation, including inflectional morphology, underspecification phonology, non-concatenative morphophonology, lexical semantics, and tone systems[1].

We presuppose that the reader of the present paper is familiar with the basic features of DATR as specified in Evans & Gazdar [1989a].

The adequacy of a lexicon representation formalism depends basically on two major factors:

- its *declarative expressiveness*: is the formalism, in principle, capable of representing the phenomena in question, and does it allow for an explicit treatment of generalisations, subgeneralisations, and exceptions?

- its range of *accessing strategies*: are there accessing strategies for all applications which presuppose a lexicon (e.g. parsing, generation, ...), and do they support the development, maintenance, and evaluation of lexica in an adequate manner?

Most of the previous work on DATR has focussed on the former set of criteria, i.e. the declarative features of the language, its expressive capabilities, and its adequacy for the re-formulation of pre-theoretic informal linguistic concepts. This paper is mainly concerned with the latter set of criteria of adequacy. However, in the case of DATR, the limited access in only one direction has led to a somewhat procedural view of the language which, in particular cases, has also had an impact on the declarative representations themselves. DATR has often been characterised as a *functional* and *deterministic* language. These features are, of course, not properties of the language itself, but rather of the language together with a particular procedural interpretation. Actually, the term *deterministic* is not applicable to a declarative language, but only makes sense if applied to a procedural language or a particular procedural interpretation of a language. The DATR interpreter/compiler systems developed so far[2] have in common that they support only one way of accessing the information represented in a DATR theory. This access strategy, which we will refer to as the *standard procedural interpretation of DATR*, closely resembles the inference rules defined in Evans & Gazdar [1989a]. Even if one considers DATR neither as a tool for parsing nor for generation tasks, but rather as a purely representational device, the one-way-only access to DATR theories turns out to be one of the major drawbacks of the model.

One of the claims stated for DATR in Evans & Gazdar [1989] is that it is computationally tractable. But for many practical purposes, including lexicon development and evaluation, it is not sufficient that there is any arbitrary accessing strategy at all, but there should be an appropriate way for accessing whatever information that is necessary for the purpose in question. This is a strong motivation for investigating alternative strategies for processing DATR representations. This paper is concerned with the *reverse query problem*, i.e. the problem how a given DATR value can be mapped onto the queries that evaluate to it. A standard query consists of a *node* and a *path*, e.g. *Sheep:<orth plur>*, and evaluates to a sequence of atoms (*value*), e.g. *sheep*. A *reverse query*, on the other hand, starts with the value, e.g. *sheep*, and queries the set of node-path pairs which evaluate to it, for instance, *Sheep:<orth sing>* and *Sheep:<orth plur>*. Our solution can be be regarded as an inversion of the *parsing-as-deduction* approach of the logic programming tradition, since we treat reverse-query theorem proving as a parsing problem. We adopt a wellknown strategy from parsing technology: we isolate the context-free "backbone" of DATR and use a modified chart-parsing algorithm for CF-PSG as a theorem prover for reverse queries.

---

*This research was partly supported by the German Federal Ministry of Research and Technology (BMfT, project VERBMOBIL) at the University of Bielefeld. I would like to thank Dafydd Gibbon for very useful comments on an earlier draft of this paper.

[1] See Cahill [1993], Gibbon [1992], Gazdar [1992], and Kilbury [1992] for recent DATR applications in these areas. An informal introduction to DATR is given in Gazdar [1990]. The standard syntax and semantics of DATR is defined in Evans & Gazdar [1989a, 1989b]. Implementation issues are discussed in Gibbon & Ahoua [1991], Jenkins [1990], and in Gibbon [1993]. Moser [1992a, 1992b, 1992c, 1992d] provides interesting insights into the formal properties of DATR (see also the DATR representations of finite state automata, different kinds of logics, register operations etc. in Evans & Gazdar [1990], and Langer [1993]). Andry et al. [1993] describe how DATR can be used in speech-oriented applications.

[2] DATR implementations have been developed by R. Evans (DATR90), D. Gibbon (DDATR, ODE), A. Sikorski (TPDATRS), J. Kilbury (QDATR), G. Drexel (YADE), M. Duda (HUBDATR), and others.

For the purposes of the present paper we will introduce a DATR notation that slightly differs from the standard notation given in Evans & Gazdar [1989] in the following respects:

- the usual DATR abbreviation conventions are spelled out

- the global environment of a DATR descriptor is explicitly represented (even if it is uninstantiated)

- each node-path pair *N:P* is associated with the set of extensional suffixes of *N:P* that are defined within the DATR theory

In standard DATR notation, what one might call a non-terminal symbol, is a node-path pair (or an abbreviation for a node-path pair). In our notation a *DATR nonterminal symbol* is an ordered set $[N, P, C, N', P']$. $N$ and $N'$ are nodes or variables ranging over nodes. $P$ and $P'$ are paths or variables ranging over paths. $C$ is the set of path suffixes of N:P.

A *DATR terminal symbol* of a theory $\theta$ is an atom that has at least one occurence in a sentence in $\theta$ where it is not an attribute, i.e. where it does not occur in a path.

The *suffix-set* w.r.t. a prefix $p$ and a set of sequences $S$ (written as $\sigma(p, S)$) is the set of the remaining suffixes of strings in $S$ which contain the prefix $p$: $\sigma(p, S) = \{s | p^\wedge s \in S\}$.

Let N:P be the left hand side of a DATR sentence of some DATR theory $\theta$. Let be $\Pi$ the set of paths occurring under node $N$ in $\theta$. The *path extension constraint* of $P$ w.r.t. $N$ and $\theta$ (written as $C(P,N,\theta)$, or simply $C$) is defined as: $C(P, N, \theta) = \sigma(P, \Pi)$.

Thus, the constraint of a path $P$ is the set of path suffixes extending $P$ of those paths that have $P$ as a prefix.
**Example**: Consider the DATR theory $\theta$:

```
N:
    <> == 0
    <a> == 1
    <a b> == 2.
```

The constraint of <> (w.r.t. N and $\theta$) is $\{<a>, <a\ b>\}$, the constraint of <a> is $\{<b>\}$, and the constraint of <a b> is $\emptyset$.

We say that a sequence $S = s_1 \ldots s_n$ ($1 \leq n$) *satisfies* a constraint $C$ iff $\{x \in C | x^\wedge X = S\} = \emptyset$ (i.e. a sequence $S$ satisfies a constraint $C$ iff there is no prefix of $S$ in $C$).

Now having defined some basic notions, we can give the rules that map standard DATR notation onto our representation:

**Mapping rules**

$$
\begin{array}{lll}
N{:}P == () & \Rightarrow & [N,P,C,N',P'] \to \varepsilon \\
N{:}P == \text{atom} & \Rightarrow & [N,P,C,N',P'] \to \text{atom} \\
N{:}P == N_2{:}P_2 & \Rightarrow & [N,P,C,N',P'] \to [N_2,P_2,C,N',P'] \\
N{:}P == N_2 & \Rightarrow & [N,P,C,N',P'] \to [N_2,P,C,N',P'] \\
N{:}P == P_2 & \Rightarrow & [N,P,C,N',P'] \to [N,P_2,C,N',P'] \\
N{:}P == "N_2{:}P_2" & \Rightarrow & [N,P,C,N',P'] \to [N_2,P_2,C,N_2,P_2] \\
N{:}P == "N_2" & \Rightarrow & [N,P,C,N',P'] \to [N_2,P',C,N_2,P'] \\
N{:}P == "P_2" & \Rightarrow & [N,P,C,N',P'] \to [N',P_2,C,N',P_2] \\
\end{array}
$$

How these mapping principles work can perhaps best be clarified by a larger example. Consider the small DATR theory, below, which we will use as an example case throughout this paper:

```
House:
    <> == Noun
    <root> == house.
Sheep:
    <> == Noun
    <root> == sheep
    <affix plur> == .
Foot:
    <> == Sheep
    <root> == foot
    <root plur> == feet.
Noun:
    <orth> == "<root>" "<affix>"
    <affix sing> ==
    <affix sing gen> == s
    <affix plur> == s.
```

The application of the mapping rules to the DATR theory above yields the following result (unstantiated variables are indicated by bold letters):

[House,<>,{<root>},**N'**,**P'**] → [Noun,<>,{<root>},**N'**,**P'**]
[House,<root>,{},**N'**,**P'**] → house
[Sheep,<>,{<root>,<affix plur>},**N'**,**P'**] →
    [Noun,<>,{<root>,<affix plur>},**N'**,**P'**]
[Sheep,<root>,∅,**N'**,**P'**] → sheep
[Sheep,<affix plur>,∅,**N'**,**P'**] → $\varepsilon$
[Foot,<>,{<root>,<root plur>},**N'**,**P'**] →
    [Sheep,<>,{<root>,<root plur>},$N'$,**P'**]
[Foot,<root>,{<plur>},**N'**,**P'**] → foot

[Foot,<root plur>,∅,**N'**,**P'**] → feet
[Noun,<orth>,∅,**N'**,**P'**] → [**N'**,<root>,∅,**N'**,<root>]
                                [**N'**,<affix>,∅,**N'**,<affix>]
[Noun,<affix sing>,{<gen>},**N'**,**P'**] → ε
[Noun,<affix sing gen>,∅,**N'**,**P'**] → s
[Noun,<affix plur>,∅,**N'**,**P'**] → s

The general aim of this (somewhat redundant) notation is to put everything that is needed for drawing inferences from a sentence (especially its global environment and possibly competing clauses at the same node) into the representation of the sentence itself. Similar internal representations are used in several DATR implementations.

## 2 Inference in DATR

Both standard inference and reverse query inference can be regarded as complex substitution operations defined for sequences of DATR terminal and non-terminal symbols which apply if particular matching criteria are satisfied. In case of DATR standard procedural semantics, a step of inference is the substitution of a DATR nonterminal by a sequence of DATR terminal and non-terminal symbols. The matching criterion applies to a given DATR query and the *left* hand sides of the sentences of the DATR theory. If the LHS of a DATR sentences satisfies the matching criterion, a modified version of the right hand side is substituted for the LHS. Since the matching criterion is such that there is at most one sentence in a DATR theory with a matching LHS, DATR standard inference is deterministic and functional. The starting point of DATR standard inference is single nonterminal and the derivation process terminates if a sequence of terminals is obtained (or if there is no LHS in the theory that satisfies the matching criterion, in which case the process of inference terminates with a failure).

In terms of DATR reverse query procedural semantics, a step of inference is the substitution of a subsequence of a given sequence of DATR terminal and non-terminal symbols by a DATR non-terminal. The matching criterion applies to the subsequence and the *right* hand sides of the sentences of the DATR theory. If the matching criterion is satisfied, a modified version of the LHS of the DATR sentence is substituted for the matching subsequence. In contrast to DATR standard inference, the matching criterion is such that there might be several DATR sentences in a given theory which satisfy it. DATR reverse query inference is hence neither functional, nor deterministic. Starting point of a reverse query is a sequence of terminals (a value). A derivation terminates, if the substitutions finally yield a single nonterminal with identical local and global environment (or if there are no matching sentences in the theory, in which case the derivation fails).

We now define the matching criteria for DATR terminal symbols, DATR nonterminal symbols and sequences of DATR symbols. These matching criteria relate extensional lemmata (i.e. already derived partial analyses) to DATR definitional sentences (i.e. "rules" that may yield a further reduction) w.r.t. a given DATR theory $\theta$.

A *terminal symbol* $t_1$ *matches* another terminal symbol $t_2$ iff $t_1 = t_2$. We also say that $t_1$ *matches* $t_2$ *with an arbitrary suffix and an empty constraint* in order to provide compatibility with the definitions for *nonterminals*, below.

1. A *nonterminal* $[N, P_1, C_1, N', P']$ *matches* another nonterminal $[N, P_2, C_2, N', P']$ *with a suffix $E$ and a constraint $C_2$* if (a) $P_2 = P_1 \wedge E$, and (b) $E$ satisfies $C_1$.

2. A *nonterminal* $[N, P_1, C_1, N', P']$ *matches* another nonterminal $[N, P_2, C_2, N', P']$ with an *empty suffix and a constraint $\sigma(P_1, C_2)$* if (a) $P_1 = P_2 \wedge E$, and (b) $E$ satisfies $C_2$.

**Example**: The non-terminal symbol $[Node, <$a b$>, \{<$c d e$>\}, N_1', P_1']$ matches $[Node, <$a b c d$>, \emptyset, N_2', P_2']$ with suffix $S = <$c d$>$ and constraint $\emptyset$.

From the definitions, given above, we can derive the matching criterion for sequences:

1. The *empty sequence matches* the empty sequence with an empty suffix and constraint $\emptyset$.

2. A *non-empty sequence* of (terminal and non-terminal) symbols $s_1' \ldots s_n'$ ($1 \leq$ n) *matches* another sequence of (terminal and non-terminal) symbols $s_1 \ldots s_n$ with suffix $E$ and constraint $C$ if
(a) for each symbol $s_i$ ($1 \leq i \leq n$): $s_i'$ matches $s_i$ with suffix $E$ and constraint $C_i$, and
(b) $C = C_1 \cup C_2 \ldots \cup C_n$.

To put it roughly, this definition requires that the symbols of the sequences match one another with the *same* (possibly empty) suffix. The resulting constraint of the sequence is the union of the constraints of the symbols.

**Example**: The string of nonterminal symbols [N1,<a>,$C_1$,N'1,P'1]   [N2,<x>,$C_2$,N'2,P'2] matches [N1,<a b>,{<c>,<d>},N'1,P'1] [N2,<x b>, {<e>},N'2,P'2] with suffix <b> and constraint {<c>, <d>, <e>}.[3]

---

[3] The matching criteria, defined above, do not cover nonterminals with *evaluable paths*, i.e. paths that include (an arbitrary number of possibly recursively embedded) nonterminals. The matching criterion for nonterminals has to be extended in order to account for statements with evaluable paths: Let be eval($\alpha, e, \theta$) a function that maps a string of DATR terminal and nonterminal symbols $\alpha = A_1 \ldots A_n$ onto a string of DATR terminals $\alpha'$ such that (a) each terminal symbol $A_i (1 \leq i \leq n)$ in $\alpha$ is mapped onto itself in $\alpha'$, and (b) each nonterminal $A_j = [N_j, P_j, C_j, N_j', P_j'] (1 \leq j \leq n)$ in $\alpha$ is mapped onto the sequence $a_j^1 \ldots a_j^m$ in $\alpha'$ such that $N_j : P_j \wedge e = a_j^1 \ldots a_j^m$ in $\theta$. '$\wedge$' refers to (recur-

# 3 The Algorithm

Metaphorically, DATR can be regarded as a formalism that exhibits a context-free backbone[4]. In analogy to a context-free phrase structure rule, a DATR sentence has a left hand side that consists of exactly one non-terminal symbol (i.e. a node-path pair) and a right hand side that consists of an arbitrary number of non-terminal and terminal symbols (i.e. DATR atoms). In contrast to context-free phrase structure grammar, DATR nonterminals are not atomic symbols, but highly structured complex objects. Additionally, DATR differs from CF-PSG in that there is not a unique start symbol but a possibly infinite set of them (i.e. the set of node-path pairs that, taken as the starting point of a query, yield a value).

Despite these differences, the basic similarity of DATR sentences and CF-PSG rules suggests that, in principle, any parsing algorithm for CF-PSGs could be a suitable starting point for constructing a reverse query algorithm for DATR. The algorithm adopted here is a *bottom-up chart parser*.

A *chart parser* is an abstract machine that performs exactly one action. This action is monotonically adding *items* to an abstract data-structure called *chart*, which might be thought of as a graph with annotated *arcs* (which are also often referred to as *edges*) or a matrix. There are basically two different kinds of items:

- *inactive* items (which represent completed analyses of substrings of the input string)

- *active* items (which represent incomplete analyses of substrings of the input string)

If one thinks of a chart in terms of a graph structure consisting of vertices connected by arcs, then an item can be defined as a triple (START, END, LABEL), where START and END are vertices connected by an arc labeled with LABEL. Active and inactive items differ with respect to the structure of the label. Inactive items are labeled with a category representing the analysis of the substring given by the START and END position. An active item is labeled with a category representing the analysis for a substring starting at START and ending at some yet unknown position X (END $\leq$ X) and a list of categories that still have to be proven to be proper analyses of a sequence of connected substrings starting at END and ending at X. For the purpose of processing DATR rather than CF-PSGs, each active item is additionally associated with a path suffix. Thus an active item has the structure:

(START,END,CAT0, $CAT_1 \ldots CAT_n$, SUFFIX)

Consider the following examples: the inactive item

(0, 1, [House,<orth sing>,{<gen>},House,**P'**])

represents the information that the substring of the input string consisting of the first symbol is the value of the query *House:<orth sing>* (with any extensional path suffix, but not *gen*) in the global environment that consists of the node *House* and some still uninstantiated path **P'**. The active item

(0,1,[Noun,<orth>,$\emptyset$,House,**P'**],
[House,<affix>,$\emptyset$,House,**P'**],$\varepsilon$)

represents the information that there is a partial analysis for a substring of the input string that starts with the first symbol and ends somewhere to the right. This substring is the value of the query Noun:<orth> within the global environment consisting of the node *House* and some uninstantiated global path **P'**, if there is a substring starting from vertex 1 that turns out to be the value of the query *House:<affix>* in the same global environment *House*:**P'**.

The general aim is to get all inactive items labeled with a start symbol (i.e. a DATR nonterminal with identical local and global environment) for the whole string which a derivable from the given grammar. There are different strategies to achieve this. The one we have adopted here is based on a chart-parsing algorithm proposed in Kay [1980].

Here is a brief description of the procedures:

- *parse* is the main procedure that scans the input, increments the pointer to the current chart position, and invokes the other procedures

- *reduce* searches the DATR theory for appropriate rules in order to achieve further reductions of inactive items

- *add-epsilon* applies epsilon productions

- *complete* combines inactive and active items

- *add-item* adds items to the chart

We will now give a more detailed description of the procedures in a pseudo-code notation (the input arguments of a procedure are given in parentheses after the procedure name). Since the only chart-modifying operation is carried out as a side effect of the procedure *add-item*, there are no output values, at all.

The procedure *parse* takes as input arguments a vertex that indicates the current chart position (in the initial state this position is 0) and the suffix of the

---

sive) DATR path extension (cf. Evans & Gazdar 1989a). Notice that $e$ has no index and thus has to be the *same* for all nonterminals $A_j$. Let $X_1 = [N, P_1, C_1, N', P']$ be a nonterminal symbol including an evaluable path $P_1$. $X_1$ *matches* $[N, P_2, C_2, N', P']$ with a suffix $E$ and a constraint $C_x$ if (a) $eval(P_1, E, \theta) = \pi$, and (b) $[N, \pi^\wedge E, C_1, N', P']$ matches $[N, P_2, C_2, N', P']$ with suffix $E$ and constraint $C_x$ (according to the matching criteria, defined above).

[4]The similarity of certain DATR sentences and context-free phrase structure rules has first been mentioned in Gibbon [1992].

input string starting at this position. As long as the remaining suffix of the input string is non-empty, *parse* calls the procedures *add-epsilon*, *reduce*, and *complete*, increments the pointer to the current chart position, and starts again with the new current vertex.

**procedure** *parse*(VERTEX, $S_1 \ldots S_n$)
*variables:*
  VERTEX, NEXT-VERTEX (integer)
  $S_1 \ldots S_n$ (string of DATR symbols)
*data:* A DATR theory $\theta$
**begin**
**if** n > 0
**then**
  NEXT-VERTEX := VERTEX + 1
  **call-proc** add-epsilon(VERTEX)
  **call-proc** reduce(VERTEX, $S_1$, NEXT-VERTEX)
  **call-proc** complete(VERTEX, $S_1$, NEXT-VERTEX)
  **call-proc** parse(NEXT-VERTEX,$S_2 \ldots S_n$)
**else** add-epsilon(VERTEX)
**end**

The procedure *add-epsilon* inserts arcs for the epsilon productions into the chart:

**procedure** *add-epsilon*(VERTEX)
*variables:* VERTEX (integer)
*data:* A DATR theory $\theta$
**begin**
**for-each** rule CAT $\to \varepsilon$ in $\theta$
  **call-proc** reduce(VERTEX, CAT, VERTEX)
  **call-proc** complete(VERTEX, CAT, VERTEX)
**end**

The procedure *reduce* takes an inactive item as the input argument and searches the DATR theory for rules that have a matching left-corner category. For each such rule found, *reduce* invokes the procedure *add-item*.

**procedure** *reduce*($V_1$,$CAT_1$,$V_2$)
*data:* A DATR theory $\theta$
**begin**
**if** is-terminal($CAT_1$)
**then**
  **for-each** rule
    $[N_0,P_0,C_0,N'_0,P'_0] \to CAT_1 \ldots CAT_n$ in $\theta$
  **call-proc** add-item($V_1$,$V_2$,$[N_0,P_0,C_0,N'_0,P'_0]$,
           $CAT_1 \ldots CAT_n$,X)
**else**
  **for-each** rule
    $[N_0,P_0,C_0,N'_0,P'_0] \to CAT_1' \ldots CAT_n$ in $\theta$
    such that $CAT_1'$ matches $CAT_1$ with suffix S
    and constraint C
  **call-proc** add-item($V_1$,$V_2$,
    $[N_0,P_0,C \cup \sigma(S,C_0),N'_0,P'_0]$, $CAT_2\ldots CAT_n$,S)
**end**

The procedure *complete* takes an inactive item as an input argument and searches the chart for active items which can be completed with it.

**procedure** *complete*($V_1$,CAT,$V_2$)
*data:* A chart CH
**begin**
**if** is-terminal(CAT)
**then for-each** active item
    ($V_0$,$V_1$,$CAT_0$,$CAT_1$ $CAT_2 \ldots CAT_n$,S) in CH
    **call-proc** add-item($V_0$,$V_2$,M,$CAT_2 \ldots CAT_n$,S)
**else for-each** active item
    ($V_0$,$V_1$,$[N_0,P_0,C_0,N'_0,P'_0]$,$CAT_1 \ldots CAT_n$,S) in CH
    such that $CAT_1$ matches CAT with constraint C
    and suffix S
    **call-proc**
    add-item($V_0$,$V_2$,$[N_0,P_0,\sigma(S,C_0)\cup$ C,
    N',P'],$CAT_2 \ldots Cat_n$,S)
**end**

The procedure *add-item* is the chart-modifying operation. It takes an active item as an input argument. If this active item has no pending categories, it is regarded as an inactive item. In this case *add-item* inserts a new chart entry for the item, provided it is not already included in the chart, and calls the procedures *reduce* and *complete*. If the item is an active item, then it is inserted into the chart, provided it is not already inside.

**procedure** *add-item*($V_1$,$V_2$,$[N_0,P_0,C_0,N'_0,P'_0]$,
    $CAT_1 \ldots CAT_n$,S)
*data:* A chart CH
**begin**
**if** $CAT_1 \ldots CAT_n = \varepsilon$
**then**
  **if** ($V_1$,$V_2$,$[N_0,P_0^\wedge S,C_0,N'_0,P'_0]$) $\in$ CH
  **then end**
  **else** CH := CH $\cup$ ($V_1$,$V_2$,$[N_0,P_0^\wedge S,C_0,N'_0,P'_0]$)
**else**
  **if**
  ($V_1$,$V_2$,$[N_0,P_0,C_0,N'_0,P'_0]$,$CAT_2 \ldots CAT_n$,S) $\in$ CH
  **then end**
  **else** CH := CH $\cup$
     ($V_1$,$V_2$,$[N_0,P_0,C_0,N'_0,P'_0]$,$CAT_2 \ldots CAT_n$,S)
**end**

## 4 Cycles

A hard problem for DATR interpreters are *cycles*, i.e. DATR statements and sets of DATR statements which involve recursive definitions such that standard inference or reverse-query inference does not necessarily terminate after a finite number of steps of inference. Here are some examples of cycles:

- *simple cycles*: N:<a> == <a>.

- *path lengthening cycles*: N:<a> == <a a>.

- *path shortening cycles*: N:<a a> == <a>.

While simple cycles have to be considered as semantically ill-formed and thus typically occur as typing errors only, both path lengthening and path shortening cycles occur quite frequently in many DATR representations. Note that path lengthening cycles turn out to be path shortening cycles in the reverse query direction, and vice versa. The DATR inference engine can be prevented from going lost in path-lengthening and path-shortening cycles by a limit on path length. This finite bound on path length can be integrated into our algorithm by modifying the *add-item* procedure such that only items with a path shorter than the permitted maximum path length are added to the chart.

## 5 Complexity

CF-PSG parsing is known to have a cubic complexity w.r.t. the length of the input string. Though it is crucial for our approach that we exploit the CF-backbone of DATR for computing reverse queries, this result is of no significance, here. DATR is Turing-equivalent (Moser 1992d), and Turing-equivalence has also been shown for a proper subset of DATR (Langer 1993). These theoretical results may a priori outrule DATR as an implementation language for large scale real time applications, but not as a development environment for prototype lexica which can be transformed into efficient task-specific on-line lexica (Andry et al. 1992). With a finite bound on path length our algorithm works, in practice[5], fast enough to be regarded as a useful tool for the development of small and medium scale lexica in DATR.

## 6 Conclusions

We have proposed an algorithm for the evaluation of reverse queries in DATR. This algorithm makes DATR-based representations applicable for various parsing tasks (e.g. morphological parsing, lexicalist syntactic parsing), and provides an important tool for lexicon development and evaluation in DATR.

## References

[Andry et al. 1992] François Andry, Norman M. Fraser, Scott McGlashan, Simon Thornton & Nick J. Youd [1992]: Making DATR Work for Speech: Lexicon Compilation in SUNDIAL. In: *Comp. Ling.* Vol. 18, No. 3, pages 245-267.

[Cahill 1993] Lynne J. Cahill: Morphonology in the Lexicon. In *Sixth Conference of the European Chapter of the Association for Computational Linguistics*, pages 87-96, 1993.

[Evans & Gazdar 1989a] Roger Evans & Gerald Gazdar: Inference in DATR. In *Fourth Conference of the European Chapter of the Association for Computational Linguistics*, pages 66-71, 1989.

[Evans & Gazdar 1989b] Roger Evans & Gerald Gazdar: The Semantics of DATR. In: Anthony G. Cohn [ed.]: *Proceedings of the Seventh Conference of the Society for the Study of Artificial Intelligence and Simulation of Behaviour*, pages 79-87, London 1989, Pitman/Morgan Kaufmann.

[Evans & Gazdar (eds.) 1990] Evans, Roger & Gerald Gazdar [eds.]: The DATR Papers. Brighton: University of Sussex Cognitive Science Research Paper CSRP 139, 1990.

[Gazdar 1992] Gerald Gazdar: Paradigm Function Morphology in DATR. In: L. J. Cahill & Richard Coates [eds.]: Sussex Papers in General and Computational Linguistics: Presented to the Linguistic Association of Great Britain Conference at Brighton Polytechnic, 6th-8th April 1992. Cognitive Science Research Paper (CSRP) No. 239. University of Sussex, 1992, pages 43-54.

[Gibbon 1992] Dafydd Gibbon: ILEX: A linguistic approach to computational lexica. In: Ursula Klenk [ed.]: Computatio Linguae. Aufsätze zur algorithmischen und quantitativen Analyse der Sprache, pages 32-53.

[Gibbon 1993] Dafydd Gibbon: Generalised DATR for flexible access: Prolog specification. English/Linguistics Occasional Papers 8. University of Bielefeld.

[Gibbon & Ahoua 1991] Dafydd Gibbon & Firmin Ahoua: DDATR: un logiciel de traitement d'héritage par défaut pour la modélisation lexical. Chiers Ivoriens de Recherche Linguistique (civl) 27. Université Nationale de Côte d'Ivoire. Abidjan, 1991, pages 5-59.

[Jenkins 1990] Elizabeth A. Jenkins: Enhancements to the Sussex Prolog DATR Implementation. In: Evans & Gazdar [eds.] [1990], pp. 41-61.

[Kay 1980] Martin Kay: Algorithm Schemata and Data Structures in Syntactic Processing. XEROX, Palo Alto.

[Kilbury 1993] James Kilbury: Paradigm-Based Derivational Morphology. In: Günther Görz [ed.]: KONVENS 92. Springer, Berlin etc. 1992, pages 159-168.

---

[5] A prolog implementation of the algorithm described in this paper is freely available as a DOS executable program. Please, contact the author for further information.

[Langer 1993] Hagen Langer: DATR without nodes and global inheritance. In: Proc. of 4. Fachtagung *Deklarative und prozedurale Aspekte der Sprachverarbeitung* der DGfS/CL, University of Hamburg, pages 71-76.

[Moser 1992a] Lionel Moser: DATR Paths as Arguments. Cognitive Science Research Paper CSRP 216, University of Sussex, Brighton.

[Moser 1992b] Lionel Moser: Lexical Constraints in DATR. Cognitive Science Research Paper CSRP 215, University of Sussex, Brighton.

[Moser 1992c] Lionel Moser: Evaluation in DATR is co-NP-Hard. Cognitive Science Research Paper CSRP 240, University of Sussex, Brighton.

[Moser 1992d] Lionel Moser: Simulating Turing Machines in DATR. Cognitive Science Research Paper CSRP 241, University of Sussex, Brighton.